# Quasi-Synthetic Riemannian Data Generation for Writer-Independent Offline Signature Verification

Elias N. Zois[(a)], Moises Diaz[(b)], Salem Said[(c)], Miguel A. Ferrer[(b)]

[a]University of West Attica, Aigaleo, Greece. [b]Instituto Universitario para el Desarrollo Tecnologico y la Innovacion en Comunicaciones. Universidad de Las Palmas de Gran Canaria, Campus de Tafira, Spain. [c]Université Grenoble-Alpes, Laboratoire Jean Kuntzmann

[(a)]ezois@uniwa.gr, [(b)]{moises.diaz, miguelangel.ferrer}@ulpgc.es, [(c)]salem.said@univ-grenoble-alpes.fr

*Abstract*— Offline handwritten signature verification remains a challenging task, particularly in writer-independent settings where models must generalize across unseen individuals. Recent developments have highlighted the advantage of geometrically inspired representations, such as covariance descriptors on Riemannian manifolds. However, past or present, handcrafted or data-driven methods usually depend on real-world signature datasets for classifier training. We introduce a quasi-synthetic data generation framework leveraging the Riemannian geometry of Symmetric Positive Definite matrices (SPD). A small set of genuine samples in the SPD space is the seed to a Riemannian Gaussian Mixture which identifies Riemannian centers as synthetic writers and variances as their properties. Riemannian Gaussian sampling on each center generates positive as well as negative synthetic SPD populations. A metric learning framework utilizes pairs of similar and dissimilar SPD points, subsequently testing it over on real-world datasets. Experiments conducted on two popular signature datasets, encompassing Western and Asian writing styles, demonstrate the efficacy of the proposed approach under both intra- and cross- dataset evaluation protocols. The results indicate that our quasi-synthetic approach achieves low error rates, highlighting the potential of generating synthetic data in Riemannian spaces for writer-independent signature verification systems.

## 1. INTRODUCTION

Handwriting, an outcome of the human movement, remains a fundamental human skill [1], playing a significant role in various forensic and security applications, despite the rise of digital technologies. Among handwriting-based authentication methods, handwritten signature verification presents a complex challenge [2], particularly in writer independent (WI-SV) operation where methods must generalize across unseen writers [3]. Typically, WI-SV models are learned (i.e. Trained and validated) on a development set [4] and then exploited (or tested) on blind datasets, following intra-lingual or cross-lingual evaluation protocols. On intra-lingual protocol, half of the writers in a dataset should be utilized for learning and the other half for testing and then, these subsets exchange roles. On cross-lingual protocol, the WI classifier is learned over an entire dataset and then tested on different datasets [5], [6]. Traditional WI-SV approaches involve a combination of "handcrafted" and/or "data-driven" techniques, with examples ranging from the dichotomy transform [7] to contemporary deep learning topologies [8].

Most of these cases require extensive real-world signature datasets for training. Also, methodologies employing synthetic data augmentation techniques have been proposed primarily with the GPDSsynthetic dataset [9] for model learning [10] or with the Signet-S [11], a popular deep-learning feature extraction module. Another line of research addresses augmentation techniques in the feature (or descriptor) domain [3], [12]. This solution is motivated by the fact that more important details can be found by analyzing extracted signature features rather than examining the signature image itself. However one may identify the absence of a probabilistic model that is both rigorously defined and tractable, thereby enabling the representation of the statistical variability of data. Recent studies have explored the potential of leveraging matrix manifolds for the purpose of mapping a signature image through its covariance descriptor to the symmetric positive definite (SPD) manifold [5], [6], [13] defined as the set of all $Y \in \mathbb{R}^{d \times d}$ real matrices: $Y \in P_d, Y^T - Y = \mathbf{0}^{d \times d}, v^T Y v > 0, \forall v \in \mathbb{R}^d - \mathbf{0}^d$. The "image to SPD matrix" encoding is achieved by its associated covariance matrix of $d$ image filters as a region descriptor. To this end, reference [14] introduced the framework for Riemannian Gaussian Distributions (RGD), while reference [15] used it for writer dependent (WD) signature verification.

In this study, we explore a Riemannian Gaussian Mixture Model (R-GMM) [14] as a novel approach to identify a number of $K$-synthetic writers accompanied by their statistical parameters, i.e. centers $\{M_{i=1:K} \in P_d\}$ and variances $\{\sigma_i\}$. The seed to R-GMM is a small population of genuine samples. The take-home message of the proposed article is about: a) modeling handwritten signature properties to covariance matrices – SPD points, b) create, under a soft clustering procedure, synthetic writers and c) duplicate new samples for improvement of the learning procedure. The synthetics $M_i$ are used to randomly draw synthetic SPD matrices of two kinds: In the first one, for each center $M_i$ we draw SPD samples, which mimic genuine samples, with $\sigma_i^G = \sigma_i$, while in the second one, we draw SPD samples with $\sigma_i^F = a \cdot \sigma_i, a > 1$ (where $a$ is a scaling factor), in order to generate synthetic forgeries of varying qualities. Finally, an SPD-related metric learning algorithm [6] is employed for offline signature verification on two popular datasets, demonstrating the effectiveness of our quasi-synthetic data generation by reducing dependence on real-world datasets.

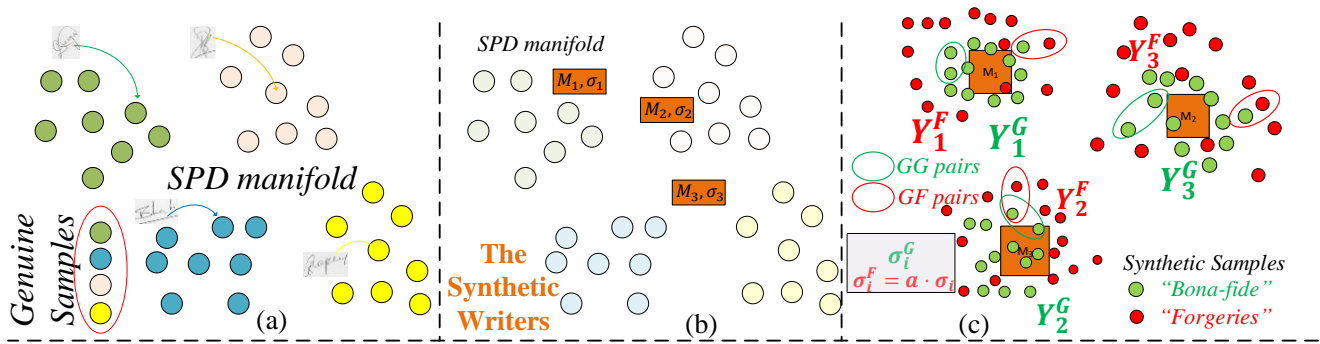

Fig. 1. Toy example of the proposed quasi-synthetic data generation. a) Out of $\mathfrak{M} = 4$ writers and $\mathfrak{D} = 8$ samples per writer we have a set of $\mathfrak{M} \cdot \mathfrak{D} = 32$ handwritten signatures and their $X = \{X_{t=1:32}\}$ SPD representations. b) A R-GMM on $X$ returns three SPD centers $M_i$ along with their corresponding variances $\sigma_i, i = 1:3$. c) For each $M_i$ we draw a number of $L = 12$ genuine $Y_i^G$ and forgery $Y_i^F$ synthetic SPD points out of two Riemannian Gaussian Distributions $Y_i^{G,F} \sim G(M_i, \sigma_i^{G,F})$ in which $\sigma_i^G = \sigma_i$, $\sigma_i^F = a \cdot \sigma_i$, $a > 1$. The metric learning based classifier will be developed using similar $Y_i^G Y_i^G$ (GG) pairs and dissimilar $Y_i^G Y_i^F$ (GF) pairs.

## 2. THE PROPOSED APPROACH

### A. The big picture

Fig. 1 presents a simple toy example in which we visually demonstrate the proposed method. Given a development set of $\mathfrak{M}$-users with $\mathfrak{D}$-development samples per user, expressed by their corresponding SPD manifold set $X = \{X_{t=1:\mathfrak{M}\cdot\mathfrak{D}}\} \in P_d$ the R-GMM algorithm [14] identifies $K$ synthetic writers and returns the model $\theta = \{\omega_i, M_i, \sigma_i\}, i = 1:K$ with $\omega_i, M_i, \sigma_i$ to be the corresponding non-zero positive weights, means and variances. Each $M_i$ center (i.e. a synthetic writer in the SPD manifold) assimilates visual attributes from the set $\{X_t\}$. Following the estimation of the $M_i$ synthetic writers we do not need anymore the human-genuine samples (a quasi-synthetic approach). Thus, synthetic genuine $Y_i^G$ and forgeries $Y_i^F$ SPD points are artificially draw from $Y_i^G \sim G(M_i, \sigma_i^G)$ and $Y_i^F \sim G(M_i, \sigma_i^F)$ RGDs. In the development stage of an WI-SV verifier, similar ($Y_i^G Y_i^G$) as well as dissimilar ($Y_i^G Y_i^F$) SPD pairs are employed in batches for representation of the positive class $\omega^+$ and the negative class $\omega^-$. Regarding the simple toy example of Fig. 1 in which we visually try to demonstrate the proposed idea, 32 are the human-genuine samples, while 36 (=12×3) are the synthetically generated samples (12 for each synthetic writer $M_i, i = 1:3$).

### B. Mixtures of RGD: Parameter estimation and generation of synthetic data

An RGD $Y \sim G(M, \sigma)$ on the SPD manifold $P_d$ resembles the typical $G(\mu, \sigma)$ in the Euclidean space $\mathbb{R}^d$. It is defined by a probability density function (p.d.f) $Y \sim G(M, \sigma)$ where $\mathbf{M} \in P_d$ is the Riemannian mean, $\sigma \in \mathbb{R}_+$ express the dispersion, $d^2(Y, \mathbf{M})$ is Rao's distance and $\zeta(\sigma)$ is a normalizing factor [14].

$$Y \sim G(M, \sigma) = \frac{1}{\zeta(\sigma)} \exp\left[-\frac{d^2(Y,M)}{2\sigma^2}\right] \quad (1)$$

Mixtures of RGDs (R-GMM) is also a probabilistic distribution on $P_d$ with p.d.f as of (2) where $K$ represents the total number of mixture components (in our case the synthetic writers) [14]:

$$p(Y) = \sum_{i=1}^{K} \omega_i \times G(M_i, \sigma_i) \quad (2)$$

Prior to the utilization of the R-GMM algorithm, we map all signature images to the corresponding covariance matrices $X_t \in P_d$; in our case $d = 10$, i.e. the manifold dimensionality is set to ten. The preprocessing steps as well as the creation of the covariance matrices $\{X_t\}$ are addressed analytically in the relative literature [5], [6]. The computation of the parameters of the R-GMM model $\theta = \{\omega_i, M_i, \sigma_i\}$ is achieved through a novel expectation maximization (EM) algorithm in the SPD manifold [14]. Any $M_i$ center along with its corresponding variance $\sigma_i$ is now employed in order to generate a population of synthetic SPD points $Y_i^G, Y_i^F$ according to (1) and Fig. 1. In our case for each center $M_i$, a total of $L = 15$ $Y_i^G \in P_{10}$ and $Y_i^F \in P_{10}$ matrices is randomly draw. To control the dispersion of synthetic $Y_i^G, Y_i^F$ points from the $M_i$ we utilize two design dispersion parameters namely: a) $\sigma_i^G = \sigma_i$ for the synthetic genuine and b) $\sigma_i^F = a \cdot \sigma_i$ for the synthetic forgeries, in which the scaling factor $a$ controls the dispersion of $Y_i^F \sim G(M_i, \sigma_i^F)$.

In this work the scaling factor $a$ was set to a number of two ad-hoc values: $a = 1.1$ or $a = 2$ in order to intuitively identify two diverse forgery qualities: in the first the generated $Y_i^F$ samples are considered to be proximal to the $Y_i^G$, while in the second case they are apart. Therefore, during the development stage of the classifier and for $a = 1.1$ dissimilar $Y^G Y^F$ pairs will tend to exhibit $d^2(Y_i^G, Y_i^F)$ distances that significantly overlap with the $d^2(Y_i^G, Y_i^G)$ distances, contrary to the $a = 2$ case. We consider the $a = 1.1$ case to correspond to the scenario in which the $\omega^-$ class is represented by pairs of real-world genuine and skilled forgeries (GSF) while the $a = 2$ case corresponds to the scenario in which the $\omega^-$ class is represented by pairs of genuine and other people's genuine (or random forgeries, GRF). For simplicity we refer to the $a = 1.1$ and the $a = 2$ cases as "Hard" or "Soft" training.

## C. The metric learning classifier

We address the WI-SV classifier as a binary SPD metric learning problem denoted as $\varDelta = \{\{W_c\}, A, \Sigma\}$ with $\{W_c\}, A, \Sigma$ its design parameters [6]. In details, for any given pair of SPD matrices $Y_i^G Y_i^G$ or $Y_i^G Y_i^F$ the metric learning evaluates a distance $\varDelta$ as a measure of (dis)similarity between $Y_i^G Y_i^G$ or $Y_i^G Y_i^F$. Ideally, $\varDelta$ should be low for similar $Y_i^G Y_i^G$ and high for dissimilar $Y_i^G Y_i^F$ pairs. Some details regarding the learnable model parameters follow:

- $W_c \in \mathbb{R}^{d \times q}$ with $c = 1:m$ is a set of orthogonal point-to-set transformation matrices which projects any high dimensional SPD point $Y_i, \in P_d$ to a set of $m$, low dimensional SPD matrices $\{Y_i^c\} \in P_q, d = m \cdot q$ by the congruence $Y_i^c = W_c^T Y_i W_c$. Thus pairs of high dimensional SPD points $Y_i^G, Y_i^{G,F} \in P_d$ are converted to a total of $m$ sub-SPD pairs $Y_i^{G,\{c\}}, Y_i^{G,F,\{c\}} \in P_q$.

- $A \in \mathbb{R}^{m \times 2}$ is a matrix comprised by a set of $(\alpha_c, \beta_c)_{c=1}^m$ values. They are the learnable parameters of the alpha-beta divergence on each low dimensional manifold point $Y_i^c \in P_q$. For a $Y_i^G, Y_i^{G,F}$ pair on $P_d$, an arbitrary $Y_i^{G,c}, Y_i^{G,F,c}$ pair on $P_q$ and two scalars $(\alpha_c, \beta_c)$ the $d^{(\alpha_c,\beta_c)}(Y_i^{G,c} \| Y_i^{G,F,c})$ is a positive real number according to:

$$d_c^{(\alpha_c,\beta_c)}(Y_i^{G,c} \| Y_i^{G,F,c}) = \frac{1}{\alpha_c \cdot \beta_c} \log\left(\det\left(\frac{\alpha_c + \beta_c \left(Y_i^{G,c}(Y_i^{G,F,c})^{-1}\right)^{-\alpha_c}}{\alpha_c + \beta_c}\right)\right) \quad (3)$$

- $\Sigma \in P_m$ is an SPD matrix entailed to map the embedding vector $v = \{d_c^{(\alpha_c, \beta_c)}\} \in \mathbb{R}^m$ to the final $\Delta \in \mathbb{R}$ distance: $\Delta = v^T \Sigma v$. For simplicity reasons, this work employ the $q = 10, m = 1$ values. Thus, $c = 1, A \in \mathbb{R}^2$ and therefore $\Sigma$ is obsolete.

### 3. EXPERIMENTAL SETUP AND RESULTS

We demonstrate the effectiveness of the proposed method for the CEDAR (55 writers, 24 genuine and skilled forgeries for each writer) and the Bengali subset of the BHSig260 database (100 writers, 24 genuine and 30 skilled forgeries for each writer). We design an intra-lingual $F_{intra}$ and a cross-lingual $F_{cross}$ experiment in which the development and the testing sets were created as follows [5], [6] [13]: Intra-lingual ($F_{intra}$): The classifier is developed on 50% of the writers and tested on the other 50%. Specifically, CEDAR: 27 writers for development, 28 for testing. Roles were reversed for a second evaluation. Bengali: 50 writers for development, 50 for testing. Roles were reversed similarly. Experiments were repeated 5 times (i.e. a 5×2 fold). Cross-lingual ($F_{cross}$): The classifier is developed on the 100% of the writers of one dataset and tested/transferred on the 100% of the other.

For each development writer we select a number of $\mathfrak{D} = 7$ genuine samples. Then, the R-GMM of (2) is applied for $K = 3, 7, 11, 15$ centers. Next, for each center we generate $L = 15$ synthetic samples according to the procedure described in section II and Fig. 1 for $Y_i^G$ and $Y_i^F, i = 1:K$ and for both cases of the scaling factor $a = 1.1$ (Hard) or $a = 2$ (Soft). Then, pairs of $Y_i^G Y_i^G, Y_i^G Y_i^F$ are selected in order to learn and validate the $\Delta = \{\{W_c\}, A, \Sigma\}$ distance. The testing stage of [5], [6] [13], visually depicted in Fig. 2, is repeated ten times according to the following procedure:

- For each writer of the testing set ten genuine samples are reserved as the reference image set $\{\mathcal{R}\}$ and SPD points $\{\mathcal{R}\}$. The remaining genuine ($\omega^+$) and skilled forgeries ($\omega^-$) form the questioned SPD set $Q^\pm$ of size $|Q^\pm|$. For each $\mathcal{R}, Q$ images an array of fourteen covariance matrices $\mathcal{R}_g, Q_g, g = 1:14$ is created as a result of a spatial equimass pyramid image partition. In detail, $\mathcal{R}_1$ points to the SPD matrix of the whole image, $\mathcal{R}_2, \dots, \mathcal{R}_5$ points to the four SPD matrices of a 2×2 equimass image segmentation and $\mathcal{R}_6, \dots, \mathcal{R}_{14}$ points to the nine SPD matrices of a 3×3 equimass image segmentation. Thus, for a pair of SPD points $(\mathcal{R}, Q^\pm)$ fourteen SPD pairs $(\mathcal{R}_g, Q_g^\pm), g = 1:14$ are used as the input to the $\Delta$ distance; consequently this forms a corresponding vector of scores $\Delta_\pm \in \mathbb{R}^{14}$.

- The resulted $\Delta_\pm$ vector derived by a $(\mathcal{R}, Q^\pm)$ pair is further sorted (i.e. $s\Delta_\pm$) and next re-arranged as $\Delta_{u\pm} \in \mathbb{R}^{14}$: $\Delta_{u\pm}(g) = mean(s\Delta_\pm(1:g))$. Consequently, for all questioned and reference pairs $(\{\mathcal{R}\}, Q^\pm)$, a tensor $T_{Q^\pm} \in \mathbb{R}^{|Q^\pm| \times 10 \times 14}$ accumulates the total scores. By taking the minimum over all ten references, a final table of scores $T_{sc_{Q^\pm}} \in \mathbb{R}^{|Q^\pm| \times 14}$ is formed. We explore each $g$-column of $T_{sc_{Q^\pm}}$ by a Receiver Operating

Characteristic curve in order to evaluate the equal error rate ($EER_{SF}$ %) per each testing writer, and per the parameter $g$. For each writer this is repeated ten times; averages are reported.

Fig. 3 display the average $EER_{SF}$ (%), plotted as a function of the parameter $g$. The results indicate that the proposed quasi-synthetic approach achieves low verification error rates, demonstrating its effectiveness in reducing reliance on real-world datasets. Beside this, two key findings emerge from the comparison of the $EER_{SF}$ curves. First, the results show that the "hard" training protocol consistently outperforms the "soft" training protocol across both datasets and for both intra-dataset ($F_{intra}$) and cross-dataset ($F_{cross}$) evaluation protocols. Within the "hard" training paradigm, the ($F_{cross}$) protocol demonstrates slightly better performance than the ($F_{intra}$) one. Second, regarding the number of clusters, Fig. 2 suggests that K=7 is a favorable choice. However, the overall system performance appears largely independent of $K$, as the error curves remain relatively stable across different values of it. A slight variation is observed when training on 100% of the CEDAR and testing on 100% of the Bengali, where lower values of yield slightly better results. In summary, our approach achieves competitive results, with EER values consistently below 0.4% across all studied cases. These promising outcomes encourage further exploration of additional configurations to validate the stability of our findings.

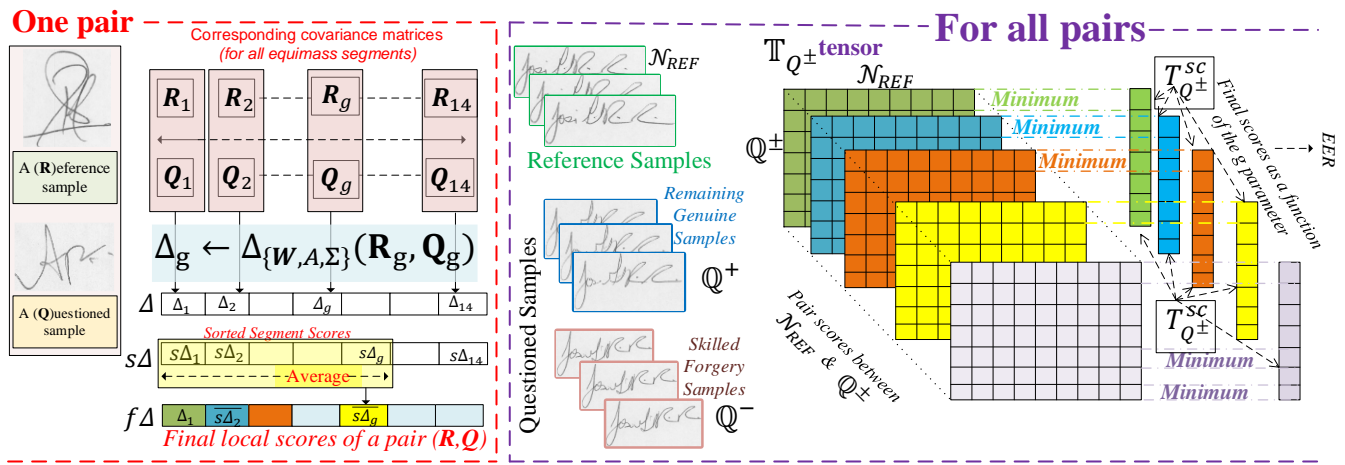

Fig. 2. The testing protocol. **Left:** Evaluation of the fourteen dimensional $f\Delta$ score vector out of one pair of signatures. **Right:** Evaluation of the EER (%) emerging from a set of reference and questioned samples.

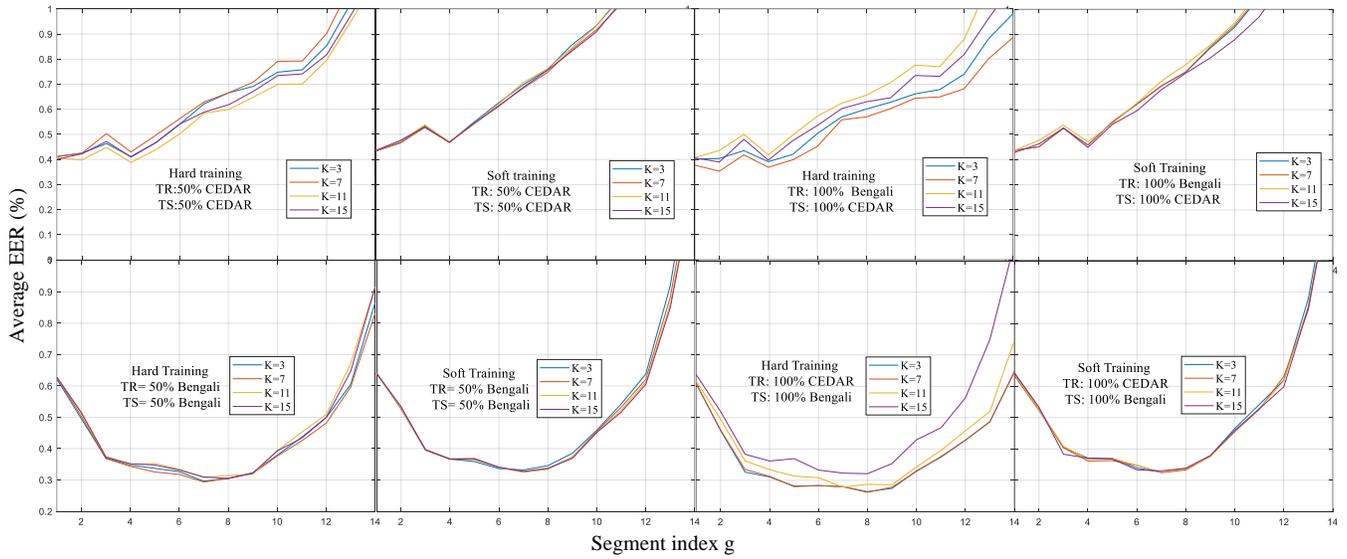

Fig.3. Average $EER_{SF}$ (%) as a function of the index $g$ for: the $F_{intra}$, $F_{cross}$, "Hard" & "Soft", protocols $L = 15$, and $K = 3, 7, 11, 15$ centers. TR, TS denotes Training/Testing datasets.

## 4. Conclusions

This work introduces a Riemannian quasi-synthetic data generation framework for writer-independent offline signature verification. Through Riemannian Gaussian Mixture models, we synthesize SPD representations that reduce dependency on extensive real-world datasets while maintain competitive verification performances. Our findings highlight the potential of leveraging synthetic data generation techniques in advancing robust and scalable signature verification systems.

### Acknowledgment

This research was partly supported by the PID2023-146620OB-I00, funded by MICIU/AEI 0.13039/501100011033 and the European Union's FEDER program, partly by the CajaCanaria and la Caixa (2023DIG05) and partly by the University of West Attica, Greece.